%% file: 00-main.tex
\documentclass[runningheads]{llncs}

\usepackage{graphicx}
\usepackage[dvipsnames]{xcolor}
\usepackage{amsmath, amssymb, amsfonts}
\usepackage{xspace}

\usepackage{lineno}
\usepackage{multirow}

\makeatletter
\DeclareRobustCommand\onedot{\futurelet\@let@token\@onedot}
\def\@onedot{\ifx\@let@token.\else.\null\fi\xspace}

\newcommand{\bR}{\ensuremath \mathbb{R}}
\newcommand{\bp}{\ensuremath \mathbf{p}}
\newcommand{\bx}{\ensuremath \mathbf{x}}
\newcommand{\by}{\ensuremath \mathbf{y}}
\newcommand{\bfeat}{\ensuremath \mathbf{f}}
\DeclareMathOperator*{\softmax}{softmax}

\makeatother

\begin{document}

\title{Geometric Attention for Prediction of Differential Properties in 3D Point Clouds}

\titlerunning{Geometric Attention for 3D Differential Properties}

\author{
Albert Matveev\inst{1} \and
Alexey Artemov\inst{1} \and
Denis Zorin\inst{2,1} \and
Evgeny Burnaev\inst{1}}

\authorrunning{A. Matveev et al.}

\institute{Skolkovo Institute of Science and Technology, Moscow, Russia \and
New York University, New York, USA}

\maketitle          

\begin{abstract}
Estimation of differential geometric quantities in discrete 3D data representations is one of the crucial steps in the geometry processing pipeline. Specifically, estimating normals and sharp feature lines from raw point clouds helps improve meshing quality and allows us to use more precise surface reconstruction techniques. When designing a learnable approach to such problems, the main difficulty is selecting neighborhoods in a point cloud and incorporating geometric relations between the points. In this study, we present a geometric attention mechanism that can provide such properties in a learnable fashion. We establish the usefulness of the proposed technique with several experiments on the prediction of normal vectors and the extraction of feature lines.

\end{abstract}

\input{01-introduction.tex}
\input{02-related.tex}
\input{03-design.tex}
\input{04-experiments.tex}
\input{05-conclusion.tex}
\input{06-bibliography.tex}

\end{document}

%% file: 01-introduction.tex
\section{Introduction}
\label{sec:intro}

Over the past several years, the amount of 3D data has increased considerably. Scanning devices that can capture the geometry of scanned objects are becoming widely available, and the computer vision community is showing a steady growth of interest in 3D data processing techniques. A range of applications includes digital fabrication, medical imaging, oil and gas modeling, and self-driving vehicles. 

The geometry processing pipeline transforms input scan data into high-quality surface representation through multiple steps. The result's quality and robustness, set aside particular algorithms for surface reconstruction, are highly dependent on the performance of previous stages.

One of the cornerstone steps in the pipeline is estimating differential geometric properties like normal vectors, curvature, and, desirably, sharp feature lines. These properties, estimated from raw input point clouds, play a significant role in the surface reconstruction and meshing processes~\cite{kazhdan2006poisson}. A multitude of algorithms for extracting such properties have been developed, however many of them require setting parameters for each point cloud separately, or performing grid search of parameters, making the computational complexity of such tasks burdensome.

On the contrary, the area of geometric deep learning has been emerging lately, which proposes tackling geometric problems with specialized deep learning architectures. 
Geometric deep learning techniques have shown success in problems of edge and vertex classification, edge prediction in graphs~\cite{bronstein2017geometric}, graph classification with applications to mesh classification~\cite{hanocka2019meshcnn}; mesh deformation~\cite{kostrikov2018surface}; point cloud classification and segmentation~\cite{guo2019deep}. In contrast, the estimation of geometric properties of surfaces has not been studied in depth. 

A recently presented Transformer architecture~\cite{vaswani2017attention} has studied the benefits of attention mechanisms for text processing, which has been established to be capable of detecting implicit relations between words in a sentence. When defining a local region in a point cloud, it is desirable to make use of such implicit relations between points, which makes attention a promising direction of research in the context of geometric problems. Such studies have started only recently, and most of the papers are focusing on semantic (classification) problems in point cloud processing. Little work has been done to improve the understanding of a geometry of the underlying surface.

In this paper, we present a novel attention-based module for improved neighborhood selection of point clouds. We call this module \emph{Geometric Attention}. We show that it increases the quality of learnable predictions of geometric properties from sampled point cloud patches. As a qualitative result, we examine neighborhoods and argue that Geometric Attention is capable of introducing meaningful relations between points.

This paper is organized as follows. In Section \ref{sec:related}, we provide an overview of related work, with the focus on geometry-related approaches and previous attention-based studies. Section \ref{sec:design} describes details of the proposed architecture. Experimental results are presented in Section \ref{sec:experiments}, with both qualitative and quantitative results for the prediction of normal vectors and feature lines. We conclude in Section \ref{sec:conclusion} with a brief discussion.

%% file: 02-related.tex
\begin{figure}[t]
\begin{center}
\includegraphics[width=\columnwidth]{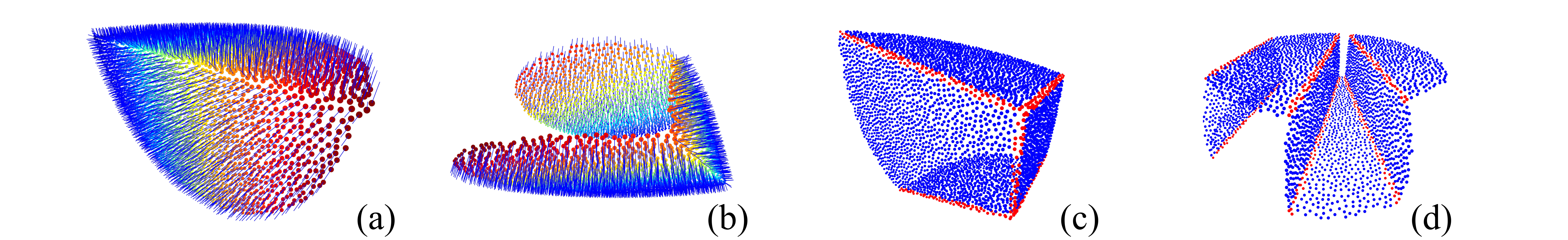}
\caption{Examples of sampled point clouds with the ground truth labels: (a), (b) -- normals, (c), (d) -- feature lines (see Section \ref{sec:experiments} for details).}
\label{fig:ground-truth-examples}
\end{center}
\end{figure}

\section{Related work}
\label{sec:related}

\textbf{Data sets.} Availability of 3D data sets has increased in recent years. Collection ShapeNet~\cite{shapenet} includes over 3 million objects. Another corpus is ModelNet~\cite{modelnet}, comprising 151,128 meshes, which is widely used for classification benchmarking. These collections of data do not fit the needs of geometric tasks due to no geometric labeling. Recently, a large-scale ABC data set~\cite{koch2019abc} has been presented. It includes over 1 million high-quality CAD models; each of them is richly annotated with geometric, topological, and semantic information. 

\textbf{Differential quantities estimation} is a standard problem for discrete surface processing. Since this problem is local, a point neighborhood is typically approximated using the $k$ nearest neighbors ($k$NN). 
The most basic types of methods rely on fitting a local surface proxy~\cite{osculatingjets}; others utilize statistical analysis techniques (e.g., ~\cite{mura}). 
A closely related property is sharp edges. This topic has not been studied in a learnable setup. Standard approaches to sharp features include analysis of covariance matrices~\cite{bazazian2015fast,merigot2010voronoi} and clustering of normals~\cite{fastrobustsharp}. Typically extracted sharp features are noisy and unstable.

\textbf{Geometric deep learning on point clouds} is a particularly popular research direction, as such architectures make minimal assumptions on input data. The primary limitation of these architectures is that they struggle to define point neighborhoods efficiently. The earlier instance of this type of networks is PointNet~\cite{pointnet} and its successor PointNet++~\cite{pointnetpp}. PointNet++ relies on the spatial proximity of points at each layer of the network, composing a point cloud's local structures based on the Euclidean nearest neighbors approach. Some work has been done to improve neighborhood query, including non-spherical $k$NN search~\cite{Sheshappanavar_2020_CVPR_Workshops}.
Similarly to these networks, Dynamic Graph CNN (DGCNN)~\cite{wang2019dynamic} utilizes Euclidean nearest neighbors as an initial neighborhood extraction; however, these local regions are recomputed deeper in the network based on learned representations of points. PointWeb~\cite{Zhao_2019_CVPR} has extended this architecture by defining a learnable point interaction inside local regions.
Other networks base their local region extraction modules on the volumetric idea by dividing the volume that encloses point cloud into grid-like cells or constructing overlapping volumes around each point~\cite{hua-pointwise-cvpr18}.

\textbf{Attention.} After the attention mechanism was shown to be beneficial in~\cite{vaswani2017attention}, many studies have adopted it for re-weighting and permutation schemes in their network architectures. In point clouds, attention has been used to refine classification and segmentation quality \cite{liu2019point2sequence,yang2019modeling}. 
Reasoning similar to ours was presented in~\cite{zhao2019rotation}. The authors decided to separate local and global information to ensure the rotational invariance of the network in the context of the semantic type of problems.

%% file: 03-design.tex
\begin{figure}[h!]
\begin{center}
\includegraphics[width=0.52\columnwidth]{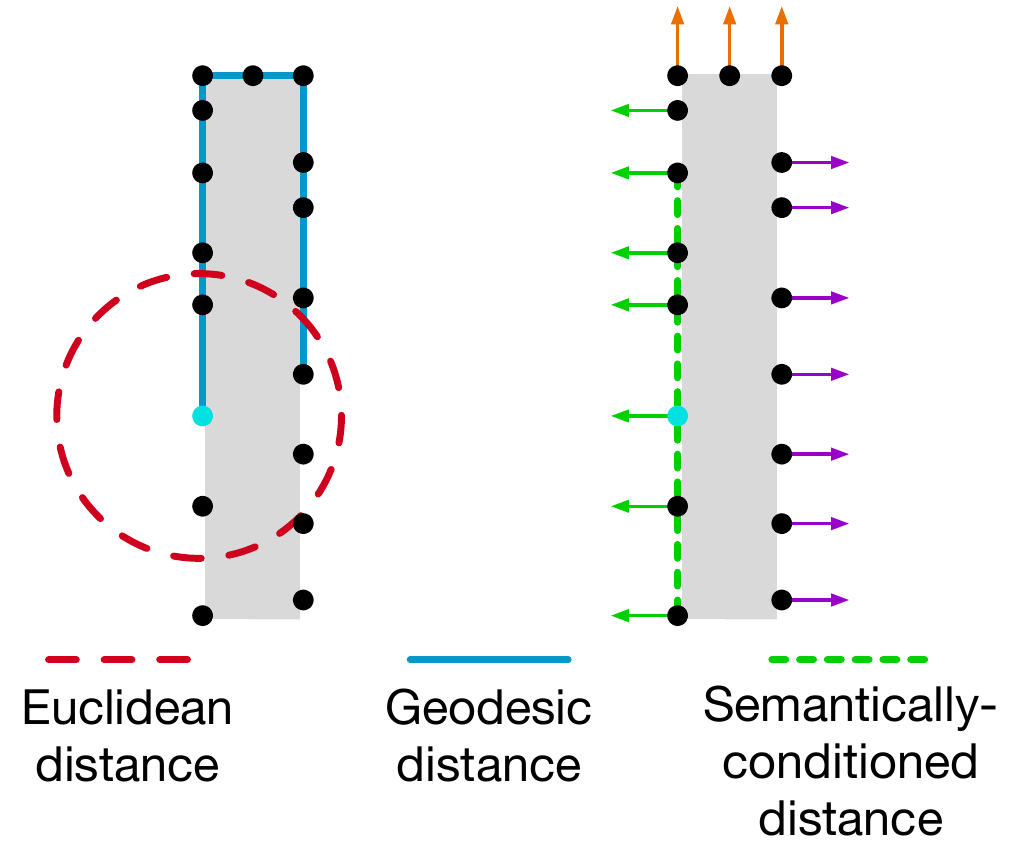}
\caption{Different types of distances.}
\label{fig:distances}
\end{center}
\end{figure}

\section{Neural network architecture}
\label{sec:design}

We start with a general problem statement; then, we describe DGCNN architecture with the main design choices. After that, we proceed with a description of the Geometric Attention module, which we incorporate into the DGCNN architecture.

Suppose you have a point cloud $P \in \bR^{N \times 3}$, consisting of $N$ points: 
$$\bp_i = (x_i, y_i, z_i) \in \bR.^3$$
The goal is to construct a mapping 
$$\varphi(\bp_i) = \by_i,$$ 
where $\by_i$ are geometric properties defined for each point $\bp_i$. The size of $\by$ may differ depending on the specific property. For instance, in case of normals vector $\by_i \in \bR^{3}$; for sharp feature labels -- $\by_i \in \{0,1\}$.

The DGCNN architecture is based on the EdgeConv operation, which, for an implicitly constructed local graph, extracts information from points as a step of propagation along edges. Technically, this is done through a proximity matrix: 
$$\mathbf{PM} = (-d_{ij}) \in \bR^{N \times N},$$
where $-d_{ij}$ is a negative distance from point $\bp_i$ to $\bp_j$.

This proximity matrix is then utilized to construct the adjacency matrix of the $k$NN graph $\mathcal{G} = \left( \mathcal{V}, \mathcal{E}\right)$ by selecting for each point $k$ nearest ones. After local areas are defined, inside each one, a multilayer perceptron (MLP) is applied to convolve neighborhood feature vectors. At the final stage, an aggregation operation (typically, a max pooling) is adopted to obtain new point features.

Following the notation from the original paper, we denote by $\bx_i \in \bR^F$ a feature vector of point $\bp_i$. Then, the EdgeConv operator is defined by
 \begin{equation}
 \label{eqn:edgeconv}
     \bx_i ^ \prime = \max_{j: (i,j) \in \mathcal E}h_{\theta}(\bx_i, \bx_j).
 \end{equation}

This EdgeConv operation is applied several times, then the outputs from each layer are concatenated together to get the final output. In such an architecture, point features carry both geometric and semantic information mixed.

\begin{figure}[h!]
\begin{center}
\includegraphics[width=0.7\columnwidth]{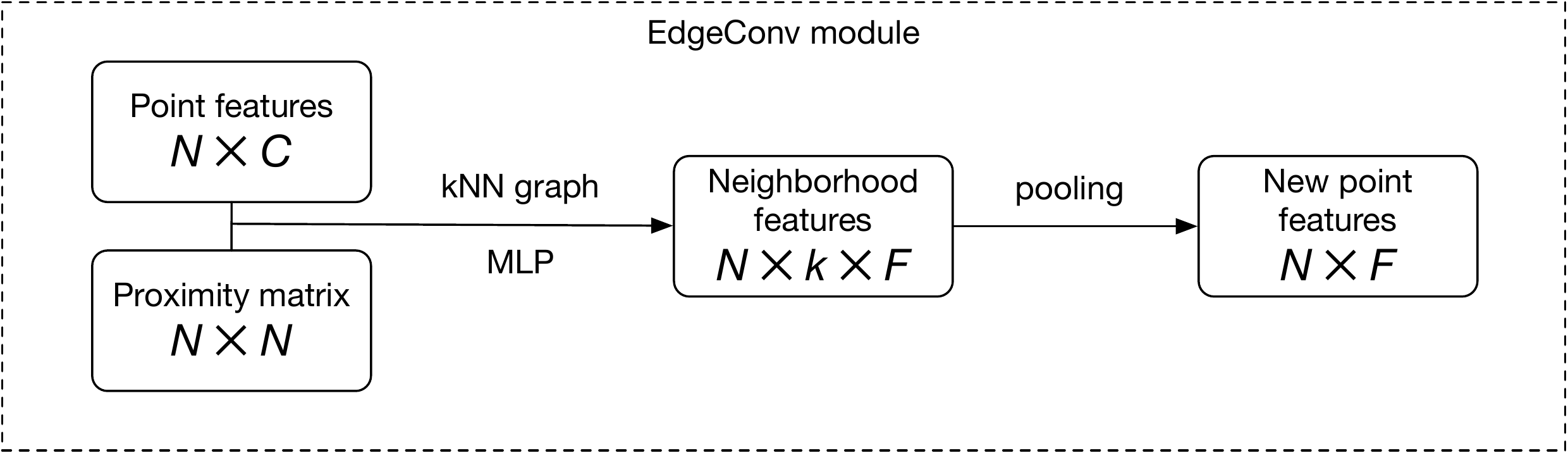}
\caption{EdgeConv module of the DGCNN architecture.}
\label{fig:edgeconv}
\end{center}
\end{figure}

\subsection{Geometric Attention module}

Our main idea is to improve the feature extraction pipeline by modifying neighborhood selection. In point cloud data, it is a common problem that sampling resolution is not enough to distinguish two sides of a thin plate (Figure \ref{fig:distances}). To solve this problem, one would require a geodesic distance defined on a point cloud, which is not easy to get. However, with additional information (normal vectors or semantic partition of a point cloud into geometric primitives), disambiguating two sides of a plate is not an issue. For this reason, we introduce a semantically-conditioned distance to represent Euclidean proximity of points if they have similar semantic features. 

\begin{figure}[h!]
\begin{center}
\includegraphics[width=0.7\columnwidth]{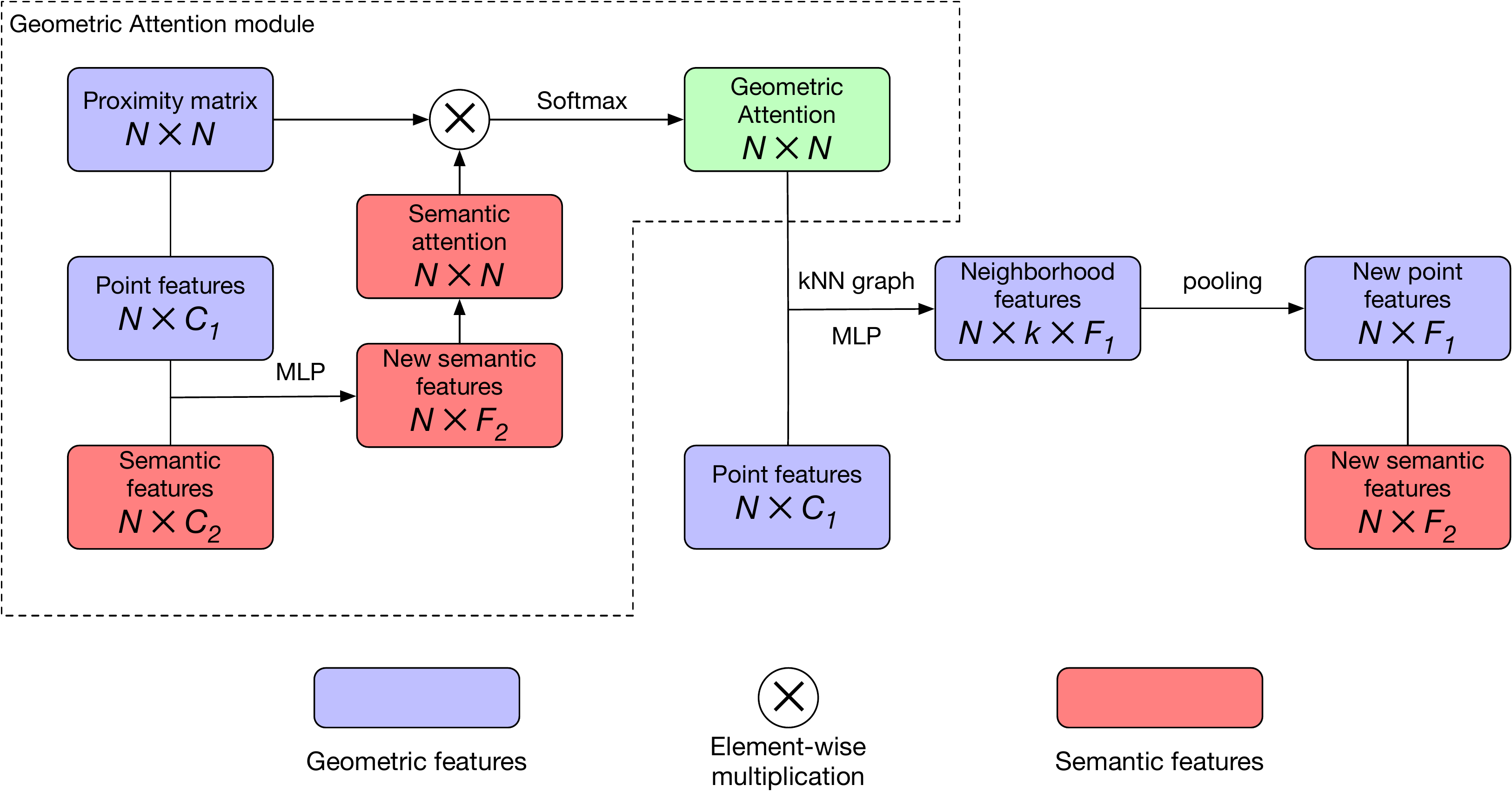}
\caption{Geometric Attention module.}
\label{fig:geomatt}
\end{center}
\end{figure}

Since such semantic information is not available, we attempt to disentangle semantic (global) features from geometric (local) information inside the network. We aim to implement this by adding semantic information flow and using both geometric proximity and semantic features-based attention in order to define the Geometric Attention module (refer to Figure \ref{fig:geomatt} for illustration).

The motivation behind this choice follows from the fact that differential quantities are defined locally, which means that the output should be computed from a small vicinity of a point. At the same time, from the global point of view, differential properties are closely related to the smoothness of a surface; they are closer to each other within one geometric primitive. For instance, a normal vector field is the same inside one planar surface patch. This ambiguity may cause a struggle for the network when computing output. Hence we attempt to have these local and global types of data separated. 

\noindent\textbf{Semantic features.} As in DGCNN, we have point features (geometric features) $\bx_i \in \bR^{F_1}$. First, we introduce semantic feature vectors $\bfeat_i \in \bR^{F_2} $, which at the first layer are separately learned from point coordinates directly. In the following layers, these features are computed from concatenated vectors of geometric features and semantic features from the previous layer. Semantic features are devoted to solely represent semantic information in a rather simple manner as a soft one-hot encoding with 64 channels. To have them represent one-hot encoding, we divide each feature vector by its norm:

\begin{equation}
 \label{eqn:semfeats}
 \begin{split}
     \bfeat_i ^ \prime &= g_{\phi}(\bx_i, \bfeat_i),\\
     \bfeat_i ^ \prime &= \frac{\bfeat_i^ \prime}{\|\bfeat_i^ \prime\|}.
 \end{split}   
\end{equation}

After semantic features are constructed, we apply Scaled Dot Product (SDP) attention to calculate the semantic attention matrix:

\begin{equation}
 \label{eqn:sematt}
 \begin{split}
    \mathbf{q}_i &= g_{\tau_1}(\bfeat_i^\prime)\\
    \mathbf{k}_i &= g_{\tau_2}(\bfeat_i^\prime)\\
     \mathbf{SA} &= \frac{\mathbf{q} \mathbf{k}^{\top}}{\sqrt{t}} = \left( \frac{\langle \mathbf{q}_i, \mathbf{k}_j \rangle}{\sqrt{t}} \right),
 \end{split}   
\end{equation}
where $t$ is a scaling factor, which is set as the dimensionality of $\bfeat_i^\prime$ according to~\cite{vaswani2017attention}.

Ideally, this leads to learning a low-rank matrix representation, with the rank equal to the number of semantic entities inside the point cloud. Since the correlation of similar semantic feature vectors is high, such a matrix should have greater values for points within the same semantic region of a point cloud.

\noindent\textbf{Semantically-conditioned proximity matrix.} Now we are ready to define the Geometric Attention matrix, or semantically-conditioned proximity matrix. Since the motivation behind DGCNN is building the graph, we follow this notion, but instead of measuring closeness of points based on $\mathbf{PM}$, we combine purely Euclidean proximity of geometric features with the learned semantic attention matrix, which encodes semantic similarity inside point cloud. To normalize these matrices, we apply the row-wise $\softmax$ function:
\begin{equation}
 \label{eqn:geomatt}
 \begin{split}
      \mathbf{SA} &= \softmax(\mathbf{SA}),\\
      \mathbf{PM} &= \softmax(\mathbf{PM}),\\
      \mathbf{GA} &= \softmax \left(\mathbf{SA} \otimes \mathbf{PM} \right)
 \end{split}   
\end{equation}
where $\mathbf{GA}$ is the Geometric Attention matrix, and $\otimes$ is element-wise matrix multiplication. The idea behind this decision is to relatively increase proximity values for those points that have similar semantic feature vectors, and decrease their closeness if semantic features are sufficiently different. 

After the matrix $\mathbf{GA}$ is computed, we follow EdgeConv as in the original paper. The rest of the architecture is structured as Dynamic Graph CNN for segmentation tasks. 

%% file: 04-experiments.tex
\section{Experimental results}
\label{sec:experiments}

We chose to predict normal vectors and detect sharp feature lines for the experimental evaluation of the Geometric Attention module. Since we focus on inferring the geometric understanding of the underlying surface, we argue that these two problems are the most representative as a benchmark. We note that the corresponding labeling could be easily obtained from the set of raw meshes, which has no additional labeling whatsoever. However, we believe that the quality of such labeling would be poor; hence, we opt to use ABC data set~\cite{koch2019abc} to simulate data for our experiments.

\subsection{Data generation and implementation details}

We start by designing the acquisition process. For a randomly selected point on a mesh surface, we begin growing the mesh neighborhood from the model by iteratively adding connected mesh faces. After the desired size has been reached, we apply the Poisson sampling technique aiming to obtain a point cloud with an average distance between points of 0.05 in original mesh units. The mesh patch size is selected such that after sampling is finished, it would ensure that the shortest sharp feature line is sampled with a predefined number of points. We found in our experiments that 8 points are sufficient to distinguish short curves robustly. When the cropped mesh patch has been sampled, we select 4096 points to provide sufficient sampling for geometric features. We refer to this set of points as point patch. We then use the initial mesh and labeling provided in the data set to transfer labels to the generated point patch. For normals, that is relatively straightforward, but for sharp features, we need to query points from point patch that are the closest to the mesh edges marked as sharp in ABC. We take point patch samples within one sampling distance tube from sharp edges and label them as "1". Using this process, we generate 200k patches and divide them into training, validation, and test sets with ratios 4:1:1, respectively. See Figure \ref{fig:ground-truth-examples} for ground truth examples.

\begin{table}[h!]
    \caption{Loss values for normals estimation and feature lines detection experiments.}
    \label{tab:table}
    \begin{center}
        \begin{tabular}{|c|p{2.7cm}p{2.7cm}|p{2.7cm}|}
            \hline
            \multirow{2}{1.5cm}{\centering Network} & \multicolumn{2}{p{5.4cm}|}{\centering Normals} & \hfil Feature lines\\
            \cline{2-4}
            & \hfil Angular loss & \hfil RMSE loss & \hfil Balanced accuracy  \\
            \hline
            DGCNN & \hfil 0.01413 & \hfil 0.38618 & \hfil 0.9753 \\
            Ours & \hfil 0.01236 & \hfil 0.38266 & \hfil 0.9892\\
            \hline
        \end{tabular}
    \end{center}
\end{table}

We normalize point patches by centering them and scaling to fit inside the unit ball. During training, we augment the training data by randomly rotating it.

Our module has been implemented using the PyTorch~\cite{paszke2019pytorch} deep learning framework. We trained networks for 10 epochs with Adam optimizer~\cite{kingma2014adam} and learning rate $10^{-3}$. Both of the networks were running with batch size 8 on one Tesla V100 GPU. We replaced all ReLU activations with LeakyReLU in order to avoid computational instabilities during normalization.

As discovered in the benchmark study devoted to normals estimation from \cite{koch2019abc}, DGCNN provides the best accuracy among learnable methods with a smaller number of parameters; hence we base our experiments on comparing to DGCNN. We note that, even though the Geometric Attention module requires additional tensors to store the features, the number of parameters does not increase considerably.

\subsection{Normals estimation}

The first task we experiment on is estimation of normal vectors. To do that, we use segmentation architecture with three output channels. We normalize the output to produce norm 1 vectors. As a loss function, we choose to optimize the loss from \cite{koch2019abc}:
\begin{equation}
 \label{eqn:loss}
     \mathcal{L} (\mathbf{n}, \mathbf{\hat{n}}) = 1 - \left( \mathbf{n} ^ \top \mathbf{\hat{n}} \right) ^ 2.
\end{equation}

\begin{figure}[t]
\begin{center}
\includegraphics[width=0.45\columnwidth]{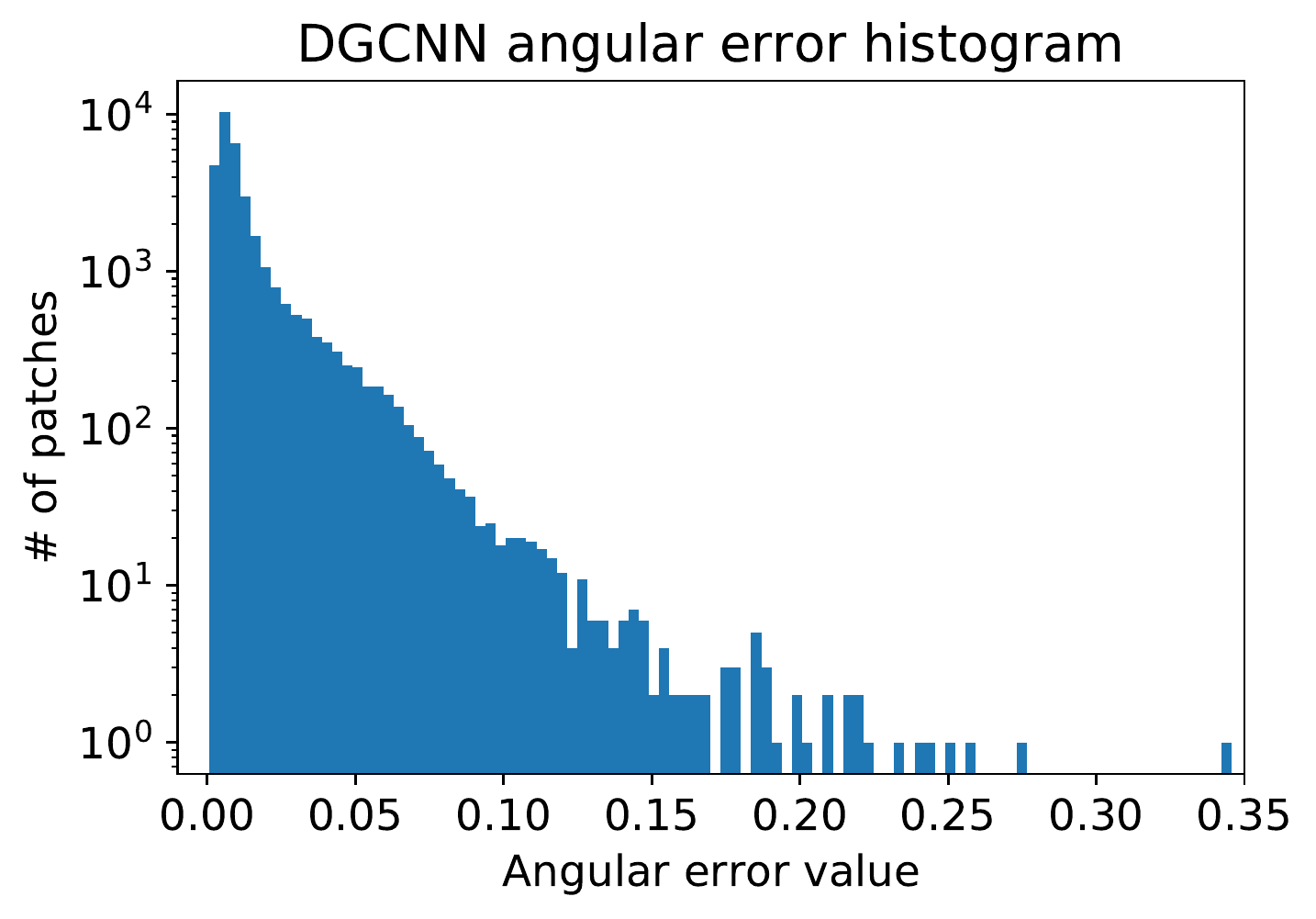}
\includegraphics[width=0.45\columnwidth]{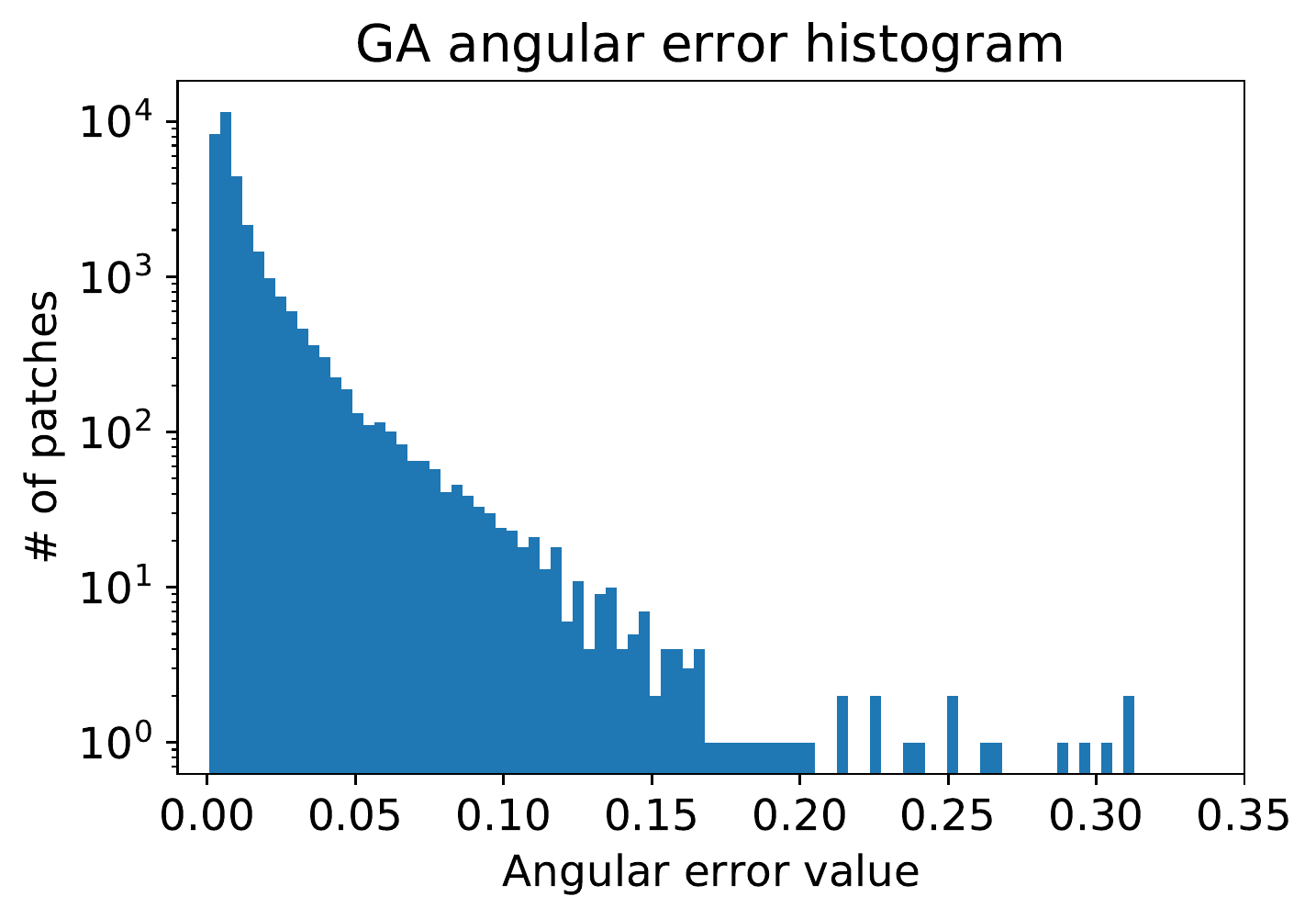}
\caption{Histogram of angular errors for normals estimation: left -- DGCNN, right -- Geometric Attention network.}
\label{fig:normals-histograms}
\end{center}
\end{figure}

Although this loss function is producing the unoriented normals, we add a small regularization with Mean Squared Error functional. Refer to Table \ref{tab:table} for numerical results, where we report the angular loss (\ref{eqn:loss}) and mean Root Mean Squared Error computed over all patches. We provide the histograms of angular errors in Figure \ref{fig:normals-histograms}. 

The histograms indicate that albeit the results are similar, the tail of the loss distribution is thinner for Geometric Attention.

\begin{figure}[h!]
\begin{center}
\includegraphics[width=0.8\columnwidth]{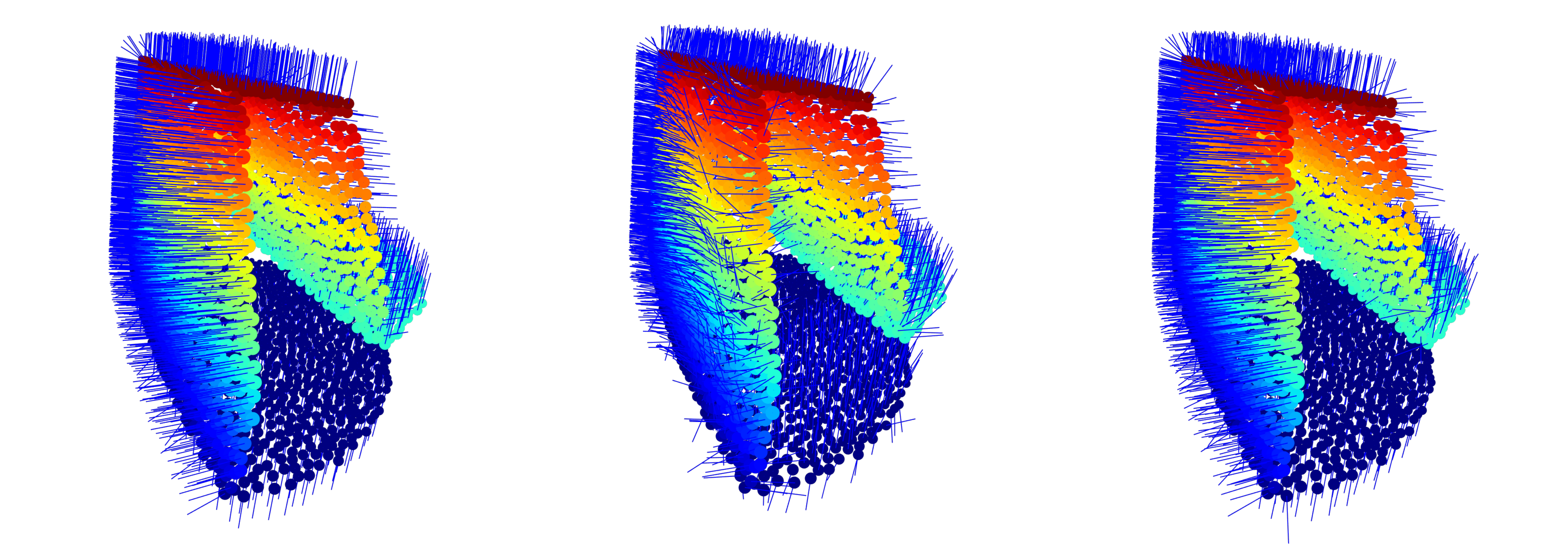}
\caption{Normals estimation result: left -- ground truth, middle -- DGCNN, right -- Geometric Attention network.}
\label{fig:normals}
\end{center}
\end{figure}

As one could see from Figure \ref{fig:normals}, DGCNN does not always take into account directions of normals, and the Geometric Attention network can determine common normals direction. We believe that semantically-conditioned distances are helping with smoothing the result while keeping it geometrically meaningful.

\subsection{Sharp feature lines extraction}

\begin{figure}[b]
\begin{center}
\includegraphics[width=0.45\columnwidth]{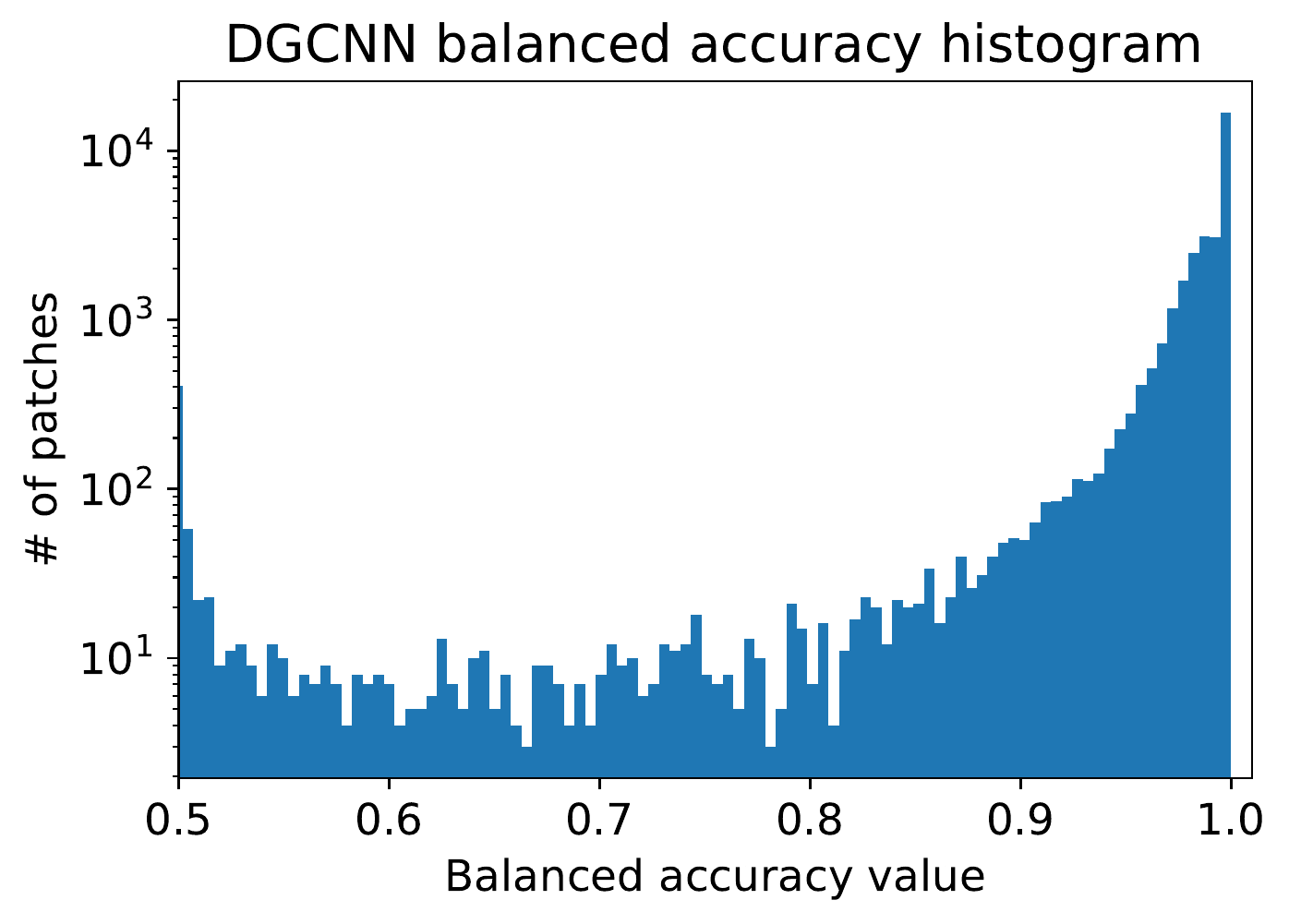}
\includegraphics[width=0.45\columnwidth]{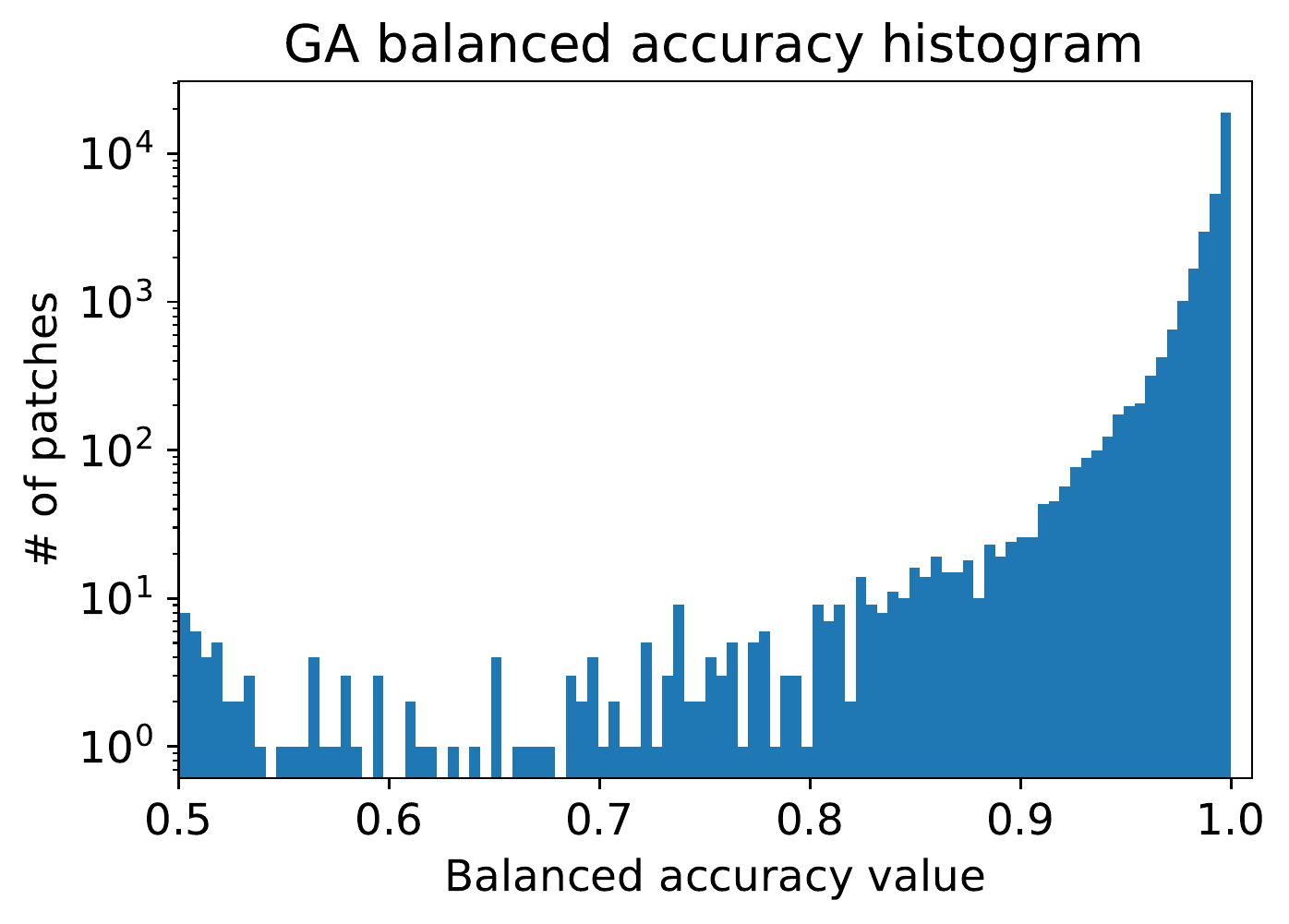}
\caption{Histogram of balanced accuracy values for sharp feature detection: left -- DGCNN, right -- Geometric Attention network.}
\label{fig:features-histograms}
\end{center}
\end{figure}

For this experiment, the segmentation architecture was made to compute one value per point. We optimized the Binary Cross-Entropy loss for this segmentation task. Table \ref{tab:table} presents the Balanced accuracy value, which was computed as an average of true negative rate and true positive rate for each patch, and then averaged over all patches. The histograms of Balanced accuracies for DGCNN and our network are in Figure \ref{fig:features-histograms}.

\begin{figure}[h!]
\begin{center}
\includegraphics[width=0.7\columnwidth]{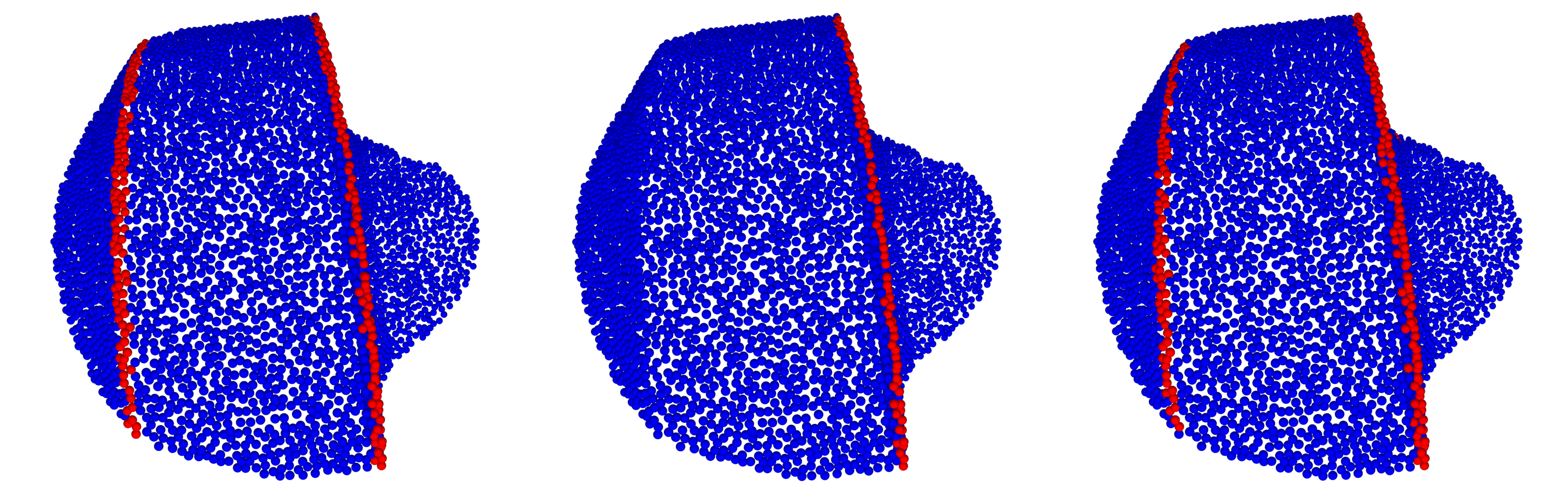}
\caption{Feature lines detection result: left -- ground truth, middle -- DGCNN, right -- Geometric Attention network.}
\label{fig:features}
\end{center}
\end{figure}

A common issue with DGCNN predictions is missing the obtuse feature lines (as seen in Figure \ref{fig:features}, middle). Our network robustly detects such cases.

\begin{figure}[b]
\begin{center}
\includegraphics[width=0.8\columnwidth]{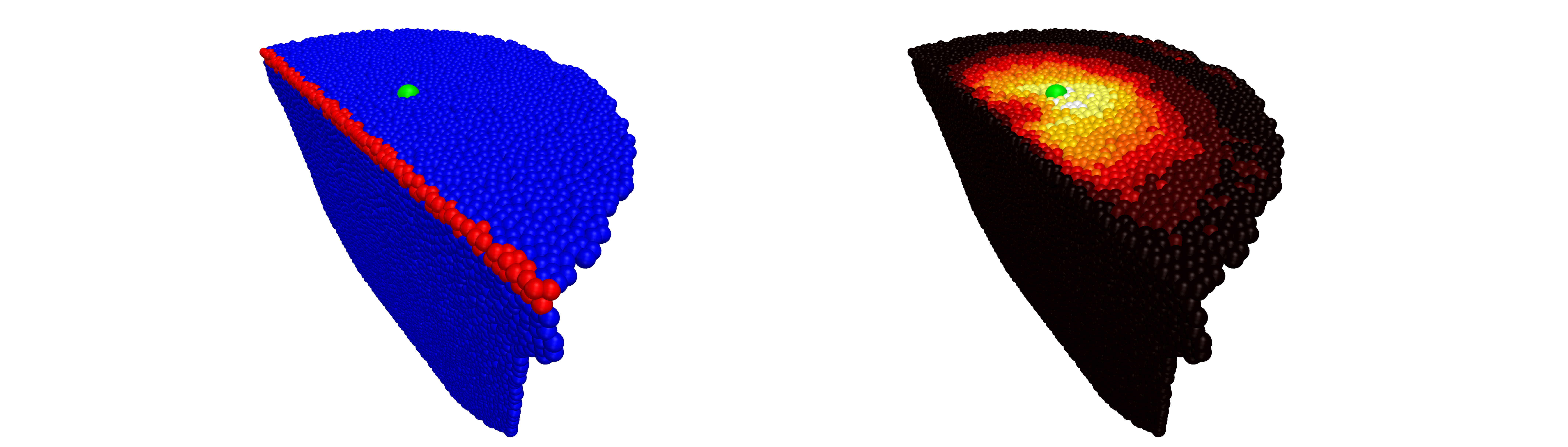}
\caption{Semantically-conditioned proximity. For a query point (large green), we show the learned distances: left -- Geometric Attention network prediction, right -- relative distances (brighter -- closer).}
\label{fig:activations}
\end{center}
\end{figure}

Lastly, we demonstrate the effect of semantically-conditioned proximity for the case of feature detection in Figure \ref{fig:activations}. It shows that the two planes have been distinguished implicitly inside the network, and the feature line semantically separates them. 
The color-coding on the right image indicates the relative distances of all points from the query point. Note that the bright region border does not extend to the set of points marked as sharp, meaning that the $k$NN would only select points from the top plane. 

%% file: 05-conclusion.tex
\section{Conclusion}
\label{sec:conclusion}

In this paper, we have proposed the Geometric Attention module, which improves point neighborhood selection in point cloud-based neural networks. Unlike the previous studies, our approach is concentrated purely on geometric properties of a point cloud.

Experiments have shown that the quality of the estimated local geometric properties of the underlying surface has increased. Qualitative results indicate that our module can meaningfully define a semantically-conditioned distance. These claims have been confirmed with two experimental setups aimed at predicting surface normals and sharp feature lines. 

The principal limitations of our approach are related to the common problems of the point cloud networks. Increasing the point patch size or the neighborhood radius leads to the rapid growth of the network, limiting scalability. The architecture design requires a fixed size of the inputs, which is not convenient in real-world applications and does not allow for the adaptive local region selection, which could be beneficial in many cases.

Possible directions of future research include the study of point interactions inside local regions for better feature extraction and further development of geometrically-inspired methods for robust geometry reconstruction from discrete surface representations.

\section*{Acknowledgments}
The reported study was funded by RFBR, project number 19-31-90144. The Authors acknowledge the usage of the Skoltech CDISE HPC cluster Zhores for obtaining the results presented in this paper.

%% file: 06-bibliography.tex
%
%
\bibliographystyle{splncs04}
\bibliography{99-bibliography.bib}
%



